\documentclass[letterpaper, 10 pt, conference]{ieeeconf}  

\IEEEoverridecommandlockouts                              
                                                          
\usepackage{cite}
\usepackage{amsmath,bm,amssymb}
\usepackage{graphicx}
\usepackage{textcomp}
\usepackage{xcolor}
\usepackage{subcaption}
\usepackage{stfloats}
\usepackage{siunitx}
\def\BibTeX{{\rm B\kern-.05em{\sc i\kern-.025em b}\kern-.08em
    T\kern-.1667em\lower.7ex\hbox{E}\kern-.125emX}}

\begin{document}

\title{\LARGE \bf Real-Time Line-Based Room Segmentation \\and Continuous Euclidean Distance Fields*
}

\author{Erik Warberg$^{1}$ \and Adam Miksits$^{1,2}$ \and Fernando S. Barbosa$^{2}$%
\thanks{*This work was partially supported by the Wallenberg AI, Autonomous Systems and Software Program (WASP) funded by the Knut and Alice Wallenberg Foundation.}%
\thanks{$^{1}$KTH Royal Institute of Technology, Sweden
        {\tt\small \{warberg, amiksits\}@kth.se}}%
\thanks{$^{2}$Ericsson Research, Sweden, 
        {\tt\small \{fernando.dos.santos.barbosa, adam.miksits\}@ericsson.com}}%
}

\maketitle

\begin{abstract}
Continuous maps representations, as opposed to traditional discrete ones such as grid maps, have been gaining traction in the research community. However, current approaches still suffer from high computation costs, making them unable to be used in large environments without sacrificing precision. In this paper, a scalable method building upon Gaussian Process-based Euclidean Distance Fields (GP-EDFs) is proposed. By leveraging structure inherent to indoor environments, namely walls and rooms, we achieve an accurate continuous map representation that is fast enough to be updated and used in real-time. This is possible thanks to a novel line-based room segmentation algorithm, enabling the creation of smaller local GP-EDFs for each room, which in turn also use line segments as its shape priors, thus representing the map more efficiently with fewer data points. We evaluate this method in simulation experiments, and make the code available open-source.
\end{abstract}


\section{Introduction}
For mobile robots, maps are essential to navigate their environment and complete their tasks. Traditionally, discrete map representations, such as occupancy grid maps, have been widely used due to their simplicity and efficiency. However, they suffer from several limitations, such as their inability to represent structures at multiple scales, susceptibility to discretization errors, and the need for significant memory resources to depict environments with sufficient detail \cite{ocallaghan_gaussian_2012}. 

The exploration of methods for constructing more accurate and memory efficient map representations of an environment has gained considerable traction. Continuous maps have shown promising results in the literature, offering a finer-grained depiction of the environment, although still at high computational costs. Furthermore, advances in image processing and machine learning algorithms have enabled better feature extraction, allowing robots to construct better abstractions of their environment, such as performing room segmentation and classification, object identification and tracking, among others.

\begin{figure}
    \centering
    \includegraphics[width=\linewidth]{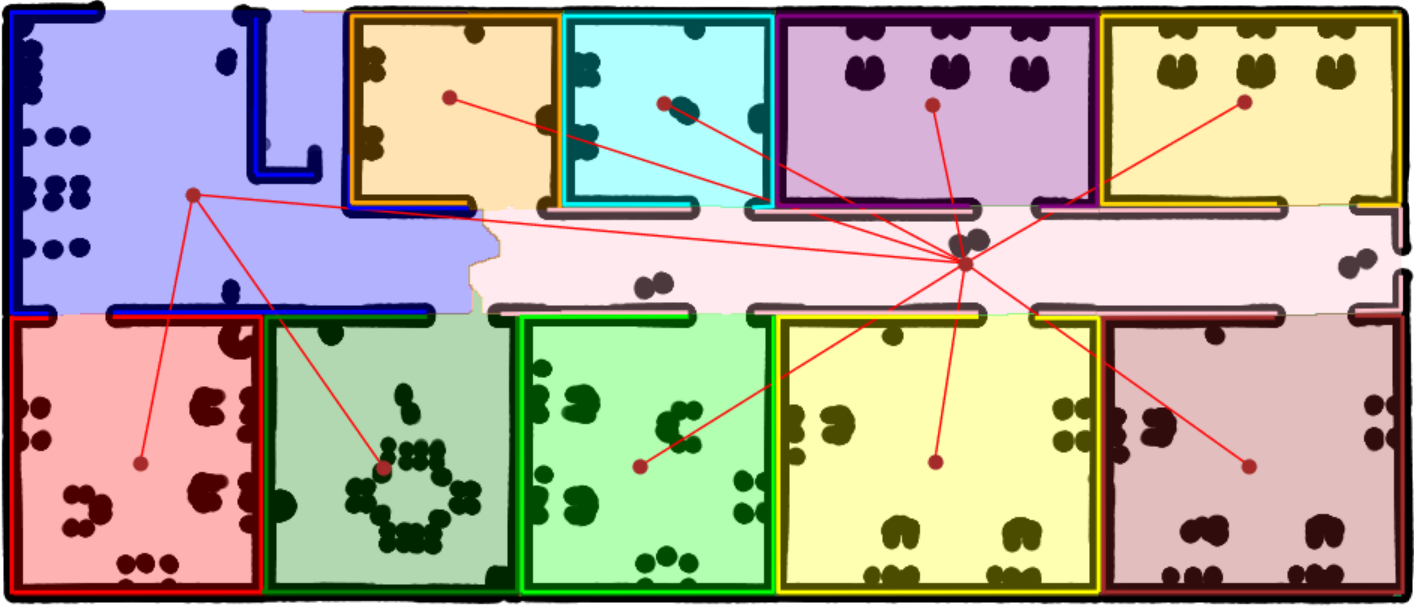}
    \caption{Result of the proposed real-time line-based room segmentation algorithm applied to a selected use case. Each room is depicted in a different color, with its line segments presented in the same color. Black dots represent obstacles detected using a 2D Lidar sensor. Dark red dots represent each room in the connectivity graph, with red lines as edges connecting neighboring rooms.}
    \label{fig:room_seg}
\end{figure}

This paper proposes leveraging line segments, extracted from measurements with a depth sensor such as a Lidar, in order to perform room segmentation and construction of continuous Euclidean distance fields (EDFs) in real time. To address the room segmentation problem, we propose a strategy for creating a visibility graph from the line segments extracted from the map, which is then used in graph clustering. Fig.~\ref{fig:room_seg} depicts the rooms segmented using our proposed method. To address the continuous EDFs problem, we use the line segments as shape priors in a specific type of Gaussian Process (GP), which is tailored for online updates. Replacing several data points in the GP dataset with line segments not only reduces the size of the dataset, but also the computational costs associated with updating and using the GP.

\subsection{Related Work}
There are numerous approaches to continuous maps, but the most prominent ones are occupancy maps and distance fields. Occupancy maps classify regions of the environment as either occupied or free-space, while distance fields provide the distance to the nearest obstacle from any point in the environment. In \cite{ocallaghan_gaussian_2012}, the authors propose using GPs to model occupancy maps by classifying the robot's environment into occupied and free space regions. Another popular method for continuous occupancy maps uses Hilbert maps, which can also capture spatial relationships between measurements naturally and possess good generalization performance \cite{ramos_hilbert_2016}. 

An advantage with distance fields as a continuous map representation is that they provide not only the distance to the nearest obstacle, but also its gradient, which can be used by optimization algorithms for collision avoidance. Distance fields can be generated with neural networks, as demonstrated in~\cite{ortiz_isdf_2022}, where a continual learning system is developed for real-time signed distance field (SDF) reconstruction, called iSDF (incremental Signed Distance Fields). Another approach to generating distance fields leverages Gaussian Process Implicit Surfaces (GPIS) to estimate either the SDF~\cite{lee_online_2019} or the EDF~\cite{wu_faithful_2021}, which enables generating the map online and using it without any post-processing.

However, GPs suffer from rapidly increasing computational costs, and numerous works have investigated how scalability can be improved without compromising accuracy. These works are usually from one of two different categories: global approximations, which use techniques such as inducing points to summarize the dataset, and local approximations, which use a divide-and-conquer approach to focus on smaller subsets of data \cite{liu_when_2020}. Global approximations are useful for capturing global patterns in the data but may not be able to capture local patterns, which is where local approximations excel. Several works have used spatial partitioning to obtain local GPs, and then either directly used the local GPs \cite{lee_online_2019, wu_faithful_2021} or combined the local models into larger models again \cite{kim_efficient_2022, stork_ensemble_2020}.

Some different approaches to room segmentation exist; in~\cite{tian_fast_2018} an approach using quadtrees and spectral clustering is presented. By gathering depth sensor data in quadtrees and exploiting some sparseness properties for clustering, rooms can be identified. In~\cite{bavle_s-graphs_2022} an algorithm is presented that instead combines free space clustering with wall plane detection to separately segment 2-wall and 4-wall rooms.

\subsection{Contribution}

The main contribution of this paper is a tightly coupled room segmentation and GP-EDF approach that exploits the structures found in indoor environments both to divide them into smaller parts, and to reduce the size of the local GP models generated for each of these parts. This enables faster model updates and inference times for even larger data sets, so that continuous maps can be used for bigger environments. The method is also made available open-source.

Our room segmentation approach is inspired by the spectral clustering approach used in~\cite{tian_fast_2018}, but to avoid clustering based on a growing set of obstacle points, we represent the walls as line segments and use these in the spectral clustering instead. Both rooms and line segments are then leveraged to generate light-weight local GP-EDFs for each room. By utilizing line shape priors as introduced in~\cite{ivan_online_2022}, each wall can be represented by one data point instead of many in the GP, which further increases efficiency by reducing the dataset size.


\subsection{Outline}
The remainder of the paper is structured as follows: Sec.~\ref{sec:room_segmentation} introduces the room segmentation approach that takes line segments as input and uses spectral clustering for room segmentation. Sec.~\ref{sec:room-based-GP-EDF} describes how individual GP-EDFs are created and maintained for each identified room. These models are then evaluated in Sec.~\ref{sec:results} in terms of computation times for updating the models and using the models for prediction. Finally, we conclude the paper in Sec.~\ref{sec:conclusions}.

\section{Room Segmentation}
\label{sec:room_segmentation}


This section first presents the novel components of the main contribution, line segment processing and room segmentation, before we proceed with explaining how they come together in the room-based GP-EDF in Section~\ref{sec:room-based-GP-EDF}. We limit the mapping to 2D, so as input, we take line segments extracted from depth sensor measurements in real-time. In our case this is accomplished by clustering data points similar to how it is done in~\cite{liu_adaptive_2022}. Line segments are then grown from clusters of depth measurements as in~\cite{gao_line_2018}, and finally merged into larger lines as in~\cite{wen_cae-rlsm_2020}. 

We represent each line segment by its two endpoints and a normal vector that points into the room. We assume to know the position of the robot, and for each line segment, the most recent position of the robot from which the line was observed is also stored. This will be used when these lines are processed to ensure distinct rooms are generated when doing spectral clustering, which is described below.

\subsection{Line Segment Processing}
\label{sec:line_segment_processing}
Before constructing the visibility graph, the endpoints of neighboring line segments are connected to generate a directed graph, which we denote $G^D$. This process involves splitting and connecting line segments, and is achieved through the three following steps: 
\subsubsection{Connecting Line Segments to Create Corners} This connects the endpoints of two non-parallel line segments. Two conditions must be met: i) the distance between the two closest endpoints of the line segments must be less than a given threshold $D_c$, and ii) the normal directions of both segments must either both point outward or both point inward of the potential corner. If the conditions are met, the coordinates of the endpoints that make the corner are modified to have the same coordinate value.
\subsubsection{Splitting Line Segments to Create Corners} This splits a line segment $l_1$ into two at the point of intersection with a non-parallel line segment $l_2$ to form a corner. To split $l_1$, the closest endpoint of $l_2$ must not be connected to any endpoint of another line segment and have a distance to $l_1$ less than the threshold $D_c$. Additionally, the intersection point must be located on $l_1$ such that it has a distance to each of its endpoints greater than the minimum allowed length of a line segment.
\subsubsection{Splitting Line Segments at Doorways} This step splits a line segment $l_1$ into two at the point of intersection with a non-parallel line segment $l_2$ where a potential doorway might be located. To split $l_1$, the closest endpoint of $l_2$ must have a distance to $l_1$ within an interval $[d_{min}, d_{max}]$ corresponding to the estimated width of a doorway. Furthermore, as in the second step, the intersection point must be located on $l_1$ such that it has a distance to each of its endpoints greater than the minimum allowed length of a line segment. If these conditions are satisfied, the closest endpoints of the two newly formed line segments are connected at the point of intersection between $l_1$ and $l_2$.  

\subsection{Visibility Graph Construction}\label{subsec:visibility_graph}

\begin{figure}
\centering
\includegraphics[width=0.7\linewidth]{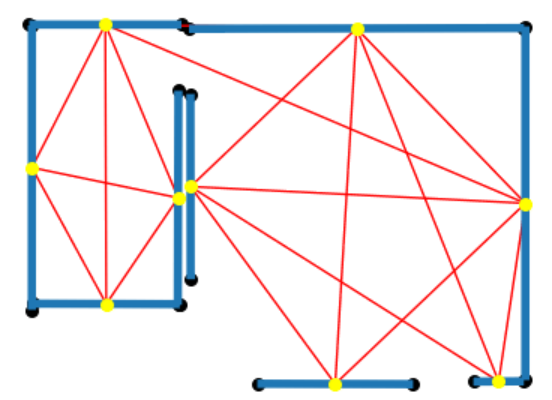}
\caption[Visibility graph visualization]{Visualization of the visibility graph, with nodes represented as blue line segments and edges as red line segments.}
\label{fig:visibility_graph}
\end{figure}

The directed graph $G^D$ is then processed into an undirected weighted visibility graph $G^V$, as depicted in Figure~\ref{fig:visibility_graph}. This process treats each line segments as a node within the graph. Edges are formed between pairs of nodes under the condition that the nodes are mutually visible or their corresponding line segments meet at their endpoints. Mutual visibility entails that the line segments do not have any other line segments obstructing their line of sight, and each of them is situated within the other's positive half-plane. Finally, different weights are assigned to the edges differently on if they meet at endpoints or if they are mutually visible.

\subsubsection*{Visibility Graph Initialization} First, $G^V$ is initialized as an undirected version of the directed graph $G^D$. At this point, the graph only contains edges between neighboring line segments, each weighted with a value of $1$ to represent the absence of distance between them. Following this, we seek pairs of line segments that are collinear, share the same direction for their normal vectors, and are separated by a gap within the range $[d_{min}, d_{max}]$. If these conditions are met, it indicates that these segments are separated by a passage. We then proceed to connect these particular line segments in the visibility graph, again assigning them a weight of $1$.

\subsubsection*{Adding Visibility Edges} Next, edges are formed between line segments based on visibility. This process involves assessing the visibility between each pair of line segments $(l_i, l_j)$ in $G^V$. For each line $l_i$, all line segments within a predefined distance threshold $D_v$ (which sets the maximum permissible distance between line segments within the same room) are identified. Of these line segments, the lines that have at least one of their endpoints in the positive half-plane of $l_i$ (as determined by its normal vector) are added to a set $\mathcal{L}^{a}$. 

Line segments that also have their midpoint on the positive half-plane of $l_i$, and a normal vector pointing in a direction that places the midpoint of $l_i$ on their positive half-plane are identified and assembled into the set $\mathcal{L}^{b}$. This set comprises potential candidates $l_j$ for forming an edge with $l_i$ in $G^V$. For each line segment $l_j$ in $\mathcal{L}^{b}$, a straight line is drawn to $l_i$ and examined for intersections with all line segments from $\mathcal{L}^{a}$. If no intersections are found, an edge is added between line segments $l_i$ and $l_j$ in $G^V$.


\subsubsection*{Computing Edge Weights} For these edges, 
a weight $W_{ij}$ is computed as a product of three weight factors:
\begin{equation}
    W_{ij} = W_{d} \cdot W_{r} \cdot W_{l}.
\end{equation}
Intuitively a higher weight indicates a greater likelihood of the lines being in the same room, taking also their respective lengths into account. The first weight factor $W_d$ is based on the shortest Euclidean distance $d(l_i, l_j)$ between $l_i$ and $l_j$, utilizing a Gaussian kernel given by the equation:
\begin{equation}
    W_{d} = \exp(-\gamma_{d} \cdot d^2(l_i,l_j)), 
\end{equation}
where $\gamma_d$ represents a tuning parameter. The second weight factor $W_r$ is calculated based on the most recent positions where the lines were observed, $\bm{x}^r_i$ and $\bm{x}^r_j$ respectively. This encodes that the two line segments are more likely to be located in the same room if their associated robot positions are close. $W_r$ is computed as:
\begin{equation}
    W_{r} = \exp(-\gamma_{r} \cdot |\bm{x}^r_i - \bm{x}^r_j|).
\end{equation}
The final weight factor $W_l$ is determined based on the lengths of the line segments, $|l_i|$ and $|l_j|$ respectively, assigning higher values to longer line segments. Normalizing with the length of the longest line segment $\max_{k \in G^V} (l_k)$ in $G^V$, the calculation of $W_l$ is given by the equation:
\begin{equation}
    W_{l} = \frac{|l_i|+|l_j|}{2 \cdot \max_{k \in G^V} (l_k)}.
\end{equation}



\subsection{Graph Clustering}
\label{sec:spectral_clustering}
After generating the visibility graph $G^V$, spectral clustering is used to partition the line segments in the graph into groups that represent distinct, meaningful rooms. The spectral clustering generally consists of the following steps \cite{von_luxburg_tutorial_2007}:

\begin{enumerate}
    \item Construct an affinity matrix $A$ where each element $A_{ij}$ corresponds to the weight of the edge between nodes $i$ and $j$ in $G^V$.
    \item Calculate the normalized symmetric Laplacian \\$L_\text{sym}=I-D^{-1/2} A D^{-1/2}$, where $D$ represents the degree matrix which carries information about the number of edges connected to each node.
    \item Perform eigenvalue decomposition of $L_\text{sym}$, sort the eigenvalues in ascending order, associate them with their corresponding eigenvectors, and select the first $k$ eigenvectors, where $k$ is the index of the largest eigengap (the difference between two consecutive eigenvalues).
    \item Perform clustering using these $k$ eigenvectors, to group the nodes of $G^V$ into $k$ separate clusters, where a clustering assignment based on a column-pivoted QR factorization, as explained in~\cite{damle_simple_2019}, is used.
\end{enumerate}

After clustering, each line segment is assigned a label based on the obtained clustering results. Since this procedure is repeated as new measurements are registered, the amount of clusters (i.e. rooms) will change over time. For instance, by denoting $k_\text{old}$ the number of clusters in the previous iteration, and $k$ that of the current iteration, if $k < k_\text{old}$, we have a situation where some rooms could be merged. In this case, the full graph is clustered again, but with the lower $k$, resulting in fewer rooms. 

However, if $k$ is larger than $k_\text{old}$, this indicates a new room might have been found. To verify this, a local visibility graph for the current room is analyzed by performing an eigenvalue decomposition on its Laplacian and examining the second smallest eigenvalue, the Fiedler value~\cite{Fiedler1973}. This value indicates graph connectivity, so if it is lower than a threshold $T_\lambda$, the local graph is clustered into two rooms. Then, the ratio of edges between the new rooms compared to the lowest number of line segments among the new rooms is computed. If the ratio is lower than a threshold $T_e$, indicating low visibility between the new rooms, the new rooms are accepted and added to the global model.

\subsection{Room Identification}\label{sec:room_identification}

Finally, the room segmentation can be used to identify in which room the robot is currently located. First, the bounding box of each detected room is computed based on their set of line segments. These bounding boxes are then subsequently inserted into an R-tree, which allows for efficient spatial lookup. Given the robots position in Cartesian space, the R-tree can thus be used to determine the robot's current room.

However, the bounding boxes of neighbouring rooms may overlap. If a position falls within overlapping bounding boxes, the R-tree will return all of the rooms corresponding to those bounding boxes. In such cases, the correct room is identified by first determining the closest line segment in each room to the query position, and then checking whether the query position falls on the positive side of each of these line segments. If the position falls on the positive side of only one line segment, then the room corresponding to that line segment is used. However, if the position falls on the positive sides of multiple line segments, the room corresponding to the closest of these line segments is selected.

\section{Room-Based GP-EDF}
\label{sec:room-based-GP-EDF}
\begin{figure*}
    \centering
    \includegraphics[width=\textwidth]{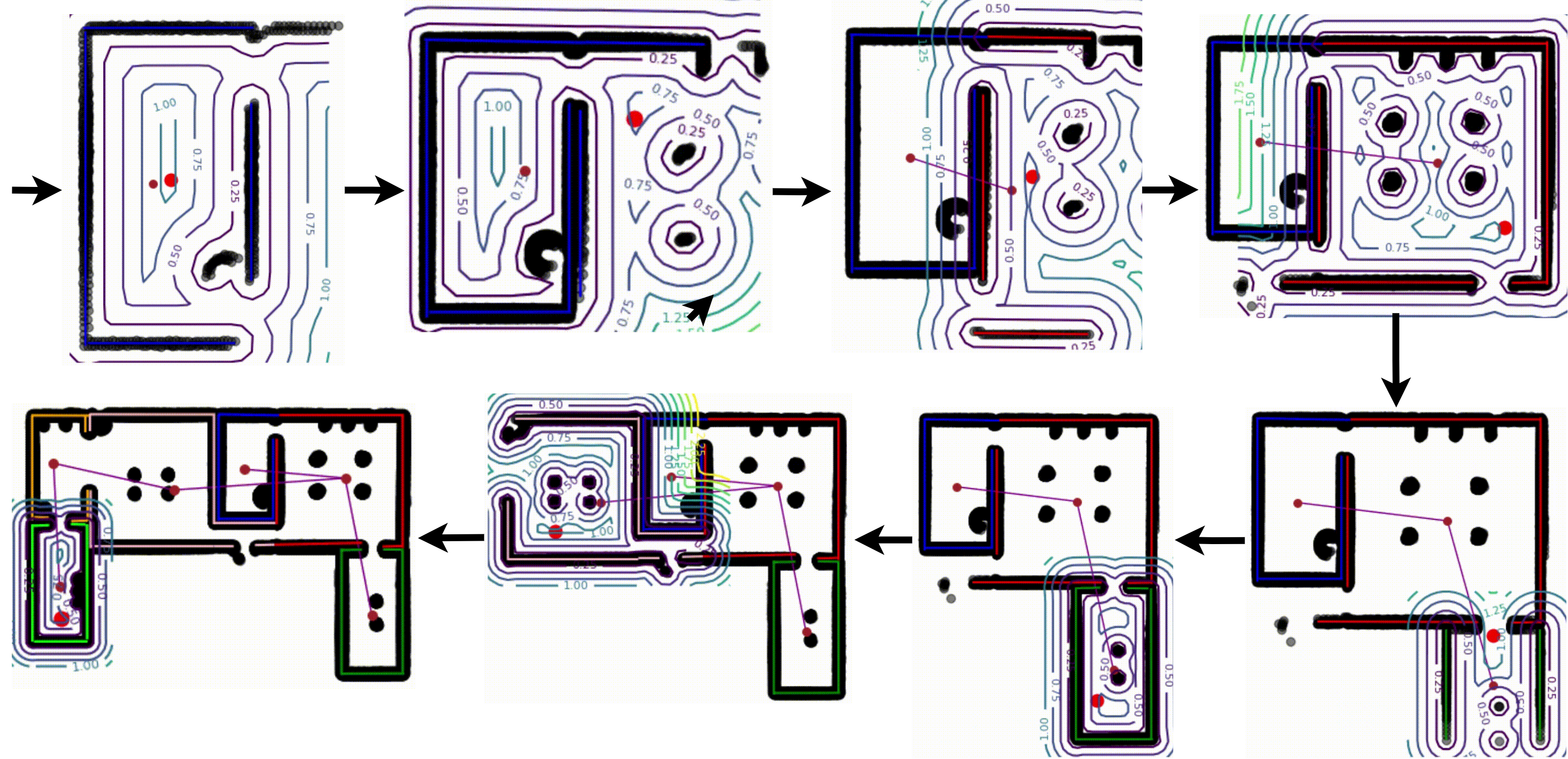}
    \caption{This figure shows a few snapshots of the room segmentation and the room-based GP-EDF being created simultaneously in real time. At the top-left corner is the very first frame after the initial sensor reading, and progress is shown clockwise. The bright red dot represents the position of the robot in the given frame, and black dots are obstacles sensed with a 2D Lidar sensor. The color of each line segment overlapping obstacle regions is represents the room to which they belong to. Colorful level curves depict the GP-EDF of the room the robot is inside of.}
    \label{fig:real-time-GP-EDF}
\end{figure*}

In this section, rooms detected in the previous section are used to generate room-based GP-EDFs with line segments as shape priors. This approach involves creating a separate, local GP-EDF model for each detected room to enable efficient updating of the models and querying of the distance to the nearest obstacle within a specific room. Fig.~\ref{fig:real-time-GP-EDF} shows snapshots of the GP-EDF being generated as a robot is navigating and detecting rooms in an environment. This process will be elaborated on before the method is evaluated in experiments.

\subsection{Line Segments as Distance Priors}
The room segmentation module provides each room with its own set of line segments, representing the walls of the rooms. These line segments, along with their simple analytic distance function, can serve as distance field priors for each GP-EDF model. By incorporating these priors, we can enhance the accuracy and computational efficiency of the GP-EDF models within each room.

Denoting the set of line segments that belong to a specific room by $\mathcal{L} = \{l_i\}_{i=1}^N$, and the distance between a query point $x$ and line segment $l_i$ as $d(x, l_i)$, one can define the line segment prior $m_\mathcal{L}(x)$ as follows:

%
\begin{equation}
    m_\mathcal{L}(x) = \exp(-\lambda \min_{i} d(x, l_i)).
\end{equation}


\subsection{Incorporating Residual Measurement Points}
In order for the GP-EDF models to capture the spatial characteristics of obstacles that cannot be represented by line segments, the measurement points that remain after the detection of line segments, the so-called residual points, are also incorporated into the local GP-EDF models. For efficiency, inducing points are adaptively selected to represent these residual points with as few points as possible.

\subsubsection*{Room Identification of Residual Points} First, each point is assigned to a single GP-EDF model. Because a sensor measurement captured in one room may overlap into neighbouring rooms through passages, both the current room and its neighbouring rooms must be analyzed to identify the precise room membership of these residual points. Thanks to clustering of sensor measurements before the line segments were detected tough, each residual point is already associated with a specific cluster, and thus a specific object. It is therefore sufficient to identify a room for each cluster of points.

To determine the exact room containing a measurement cluster, the cluster's centroid is computed by averaging all points within that cluster. Then, for each room within the sensor's range, we compute the bounding box of its corresponding set of line segments. Next, each centroid is inspected to determine which of these bounding boxes it falls inside of, and assigned a room label. If multiple or no bounding boxes are identified, a label is assigned to the centroid based on closest line segments in the same way a robot position was treated in Sec.~\ref{sec:room_identification}.

\subsubsection*{Adaptive Inducing Point Representation} To further enhance the efficiency of the GP-EDF models and avoid including excessive data points that do not contribute new information, a sparse variational approximation of the GP based on~\cite{galy-fajou_adaptive_2021} is utilized. This approximation adaptively chooses a set of inducing points $\mathcal{Z}$ over a variational distribution to represent the sensor measurement points more efficiently. The adaptive strategy asserts that a substantial proportion of sensor measurements should be located close to the set $\mathcal{Z}$ of inducing points. Intuitively, $\mathcal{Z}$ should contain a large enough amount of points, and the points should be as varied as possible to avoid redundancy and ensure a better approximation.

More specifically, for each new sensor measurement $\bm{x}$, the algorithm computes the covariance function $k$ of the Gaussian process between $\bm{x}$ and the existing set $\mathcal{Z}$ of inducing points. This results in the covariance matrix $k(\bm{x}, \mathcal{Z})$. The maximum value of $k(\bm{x}, \mathcal{Z})$ is then assessed. If this maximum value is less than a predetermined threshold $T_{\mathcal{Z}}$, the new sensor measurement $\bm{x}$ is inadequately represented by the current set $\mathcal{Z}$ of inducing points. In this instance, $\bm{x}$ is incorporated into the set $\mathcal{Z}$. Conversely, if the maximum covariance exceeds the threshold, the current set of inducing points already provides a satisfactory approximation of $\bm{x}$, and the algorithm proceeds with the next sensor measurement.

\subsection{Updating GP-EDF Models}
A GP-EDF model is initialized for the first room before any sensor measurements are obtained. As new sensor measurements arrive, new line segments are extracted and added to the mean function of the GP-EDF model corresponding to the current room of the robot. When the room segmentation module then detects a new room, the current room is divided into two parts, and the corresponding GP-EDF model must be split accordingly. Given a parent GP-EDF model $\mathcal{GP}_{p}$ with the set of inducing points $\mathcal{Z}_p$ and line segments $\mathcal{L}_p$, that is to be split into two child GP-EDF models, $\mathcal{GP}_{1}$ and $\mathcal{GP}_{2}$, the splitting process is as follows:

\begin{enumerate}
    \item Determine room membership for each line segment, as in Sec.~\ref{sec:room_identification}, and utilize this to split the line segment set $\mathcal{L}_p$ into two subsets: $\mathcal{L}_1$ corresponding to $\mathcal{GP}_1$, and $\mathcal{L}_2$ corresponding to $\mathcal{GP}_2$.
    \item Match each inducing point to one of the line segment subsets, $\mathcal{L}_1$ or $\mathcal{L}_2$. This is achieved by computing bounding boxes for each new line subset and determining which of these boxes each inducing point belongs to. If a point belongs to both or neither, the nearest line segment from each subset is identified, and the point is assigned to the line for which normal it falls on the positive side of. If both or neither of the lines pass this check, the point is matched with the closest line segment.
    
    \item Distribute the inducing points of $\mathcal{Z}_p$ and their corresponding variational distributions among the two child models. This distribution is determined by the line segment subset, either $\mathcal{L}_1$ or $\mathcal{L}_2$, to which they were matched. Consequently, $\mathcal{Z}_p$ is divided into two distinct subsets: $\mathcal{Z}_1$ and $\mathcal{Z}_2$. This division ensures $\mathcal{Z}_1 \cup \mathcal{Z}_2 = \mathcal{Z}_p$ and $\mathcal{Z}_1 \cap \mathcal{Z}_2 = \emptyset$, with each subset strictly residing within the boundaries of its corresponding room.
\end{enumerate}



Additionally, when new measurements are obtained that either introduce new line segments or extend the length of existing ones, new connections can be found between the line segments in the current robot's current room and lines in neighboring rooms. In such cases, the room segmentation process may alter the room membership of affected line segments, assigning them to another neighboring room, or even merging two neighboring rooms. This means also the GP-EDF models of the rooms have to be updated. In the first case, the affected line segments and their associated inducing points are transferred to the model of the neighboring room following the same process as for splitting. In the second case, the two models are merged by combining the sets of line segments the sets of inducing points.



\section{Results}
\label{sec:results}

In this section, the room-based GP-EDF is compared with two baseline models:
a ``Standard global model'' which constructs a single global GP-EDF for the entire environment, using a zero mean function for the GP prior, and
a ``Global line-based model'' which uses line segments as shape priors, as described in the sections above.

Simulations were conducted in a static large indoor environment with a simple structure consisting of a multitude of interconnected rooms with straight walls. Simulated sensor data was gathered by steering a mobile robot (Turtlebot3 Waffle) equipped with a 2D laser range sensor through the area using ROS2. The experiments and models were implemented using Python 3, and run on an Intel Core i7-4790 CPU with a base frequency of 3.60GHz and 16 GB of RAM.

For each of the modules of the proposed method, the values of the parameters were chosen as follows.
%
In the Room Segmentation, the line segment processing had a distance threshold of $T_c = \SI{0.4}{\meter}$ to form corners, and the interval of potential doorway widths was set to $[\SI{0.8}{\meter}, \SI{3.0}{\meter}]$. For the visibility graph, the maximum distance to form an edge was $D_v=\SI{8.0}{\meter}$, and for the edge weights the tuning parameters were chosen as $\gamma_r = 0.005$ and $\gamma_d = 0.02$. In the graph clustering, the threshold for the Fiedler value was set to $T_\lambda = 0.18$, and for re-merging rooms post-split to $T_e = 0.5$. Finally, for the GP-EDF model the lengthscale was chosen as $\lambda=100$ and the threshold for adaptive inducing point selection was $T_{cov}=1 \cdot 10^{-6}$.

\begin{figure}
     \centering
     \subfloat[Model update]{
         \includegraphics[width=0.48\textwidth]{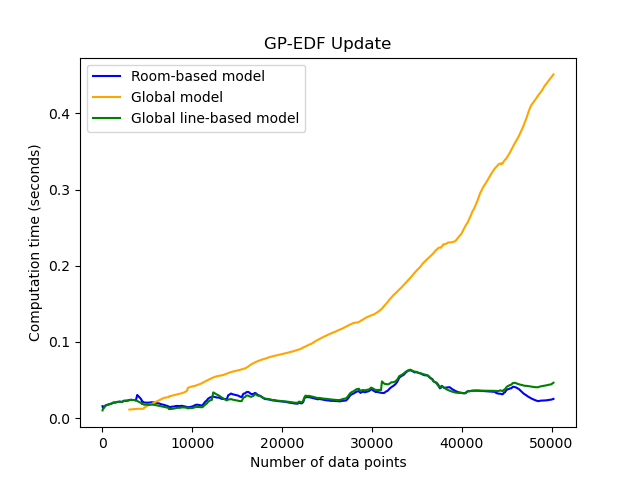}
         \label{fig:update_all_models}
         }\\
     \subfloat[Model prediction]{
         \includegraphics[width=0.48\textwidth]{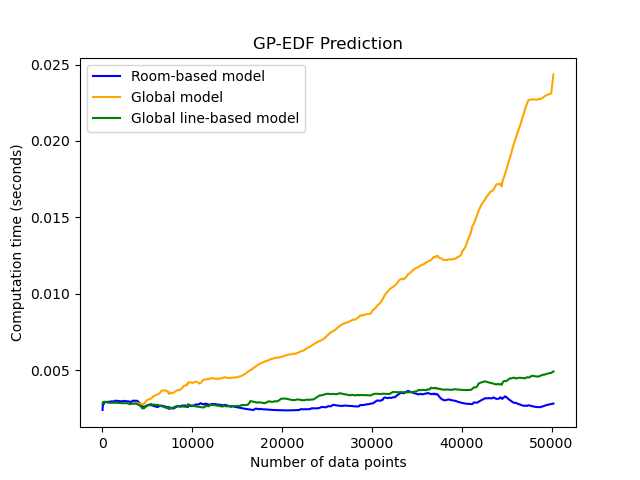}
         \label{fig:pred_all_models}
         }
     \caption[Computation time for all GP-EDF models.]{Graphs depicting changes in computation time for all GP-EDF models.}
     \label{fig:compare_time_all_models}
\end{figure}

To evaluate the required computational resources relative to the accumulated sensor data for the different GP-EDF models, the computation time was compared with the number of acquired measurement points. 
Fig.~\ref{fig:compare_time_all_models} shows the computation time in seconds in relation to the number of data points for all three methods being compared. The top plot shows the computation times for the model updates, and the bottom one for performing predictions.

For the standard global model, the computation time demonstrates an approximate cubic increase with the amount of data for both the model update and prediction processes. However, for the displayed range of data points in this figure, both the room-based model and the line-based global model maintain relatively low computation times. There is minimal discernible difference in computation time between the two models, except towards the end of the graphs.

To evaluate the computation time of each individual run of the room segmentation algorithm against the accumulation of sensor data, we analysed two different trajectories that generated partitioning similar to that in Fig.~\ref{fig:room_seg}.  
Initially, there is a notable increase in computation time (from 5ms to 15ms) as the robot starts collecting data, which can be attributed to the early stage of mapping where the creation of new rooms is still limited.
However, the computation time peaks after 50000 samples and stabilizes at approximately 18ms.

\section{Conclusions}
\label{sec:conclusions}
We proposed a method that allows for room segmentation and construction of Gaussian Process-based Euclidean distance field models in real time.
The main idea behind the method is to exploit naturally-occurring structures in indoor environments, such as walls and corridors, and abstract them away in the form of line segments.
A traditional graph clustering approach is adapted to use such line segments for improved room segmentation results.
Furthermore, the segments are employed as shape priors in a GP-EDF model.

In the future, it would be beneficial to increase the model's applicability in real-world scenarios. For instance, by relaxing the assumption on straight walls and incorporating other shapes, such as arcs and circles. Another natural extension is to a 3D model, which could have a large impact on computational requirements.
Thirdly, the potential adaptation of the model to support dynamic environments poses an interesting prospect. 
%
Finally, the assumption of perfect localization could be reassessed 
, and the method's robustness and applicability could be significantly broadened.

\section*{Acknowledgment}
The authors would like to thank Jana Tumova for her valuable feedback during this project.

\bibliographystyle{IEEEtran}
\bibliography{main} 

\begin{thebibliography}{10}
\providecommand{\url}[1]{#1}
\csname url@samestyle\endcsname
\providecommand{\newblock}{\relax}
\providecommand{\bibinfo}[2]{#2}
\providecommand{\BIBentrySTDinterwordspacing}{\spaceskip=0pt\relax}
\providecommand{\BIBentryALTinterwordstretchfactor}{4}
\providecommand{\BIBentryALTinterwordspacing}{\spaceskip=\fontdimen2\font plus
\BIBentryALTinterwordstretchfactor\fontdimen3\font minus
  \fontdimen4\font\relax}
\providecommand{\BIBforeignlanguage}[2]{{%
\expandafter\ifx\csname l@#1\endcsname\relax
\typeout{** WARNING: IEEEtran.bst: No hyphenation pattern has been}%
\typeout{** loaded for the language `#1'. Using the pattern for}%
\typeout{** the default language instead.}%
\else
\language=\csname l@#1\endcsname
\fi
#2}}
\providecommand{\BIBdecl}{\relax}
\BIBdecl

\bibitem{ocallaghan_gaussian_2012}
S.~T. O’Callaghan and F.~T. Ramos, ``Gaussian process occupancy maps,''
  \emph{The International Journal of Robotics Research}, vol.~31, no.~1, pp.
  42--62, 2012.

\bibitem{ramos_hilbert_2016}
F.~Ramos and L.~Ott, ``Hilbert maps: Scalable continuous occupancy mapping with
  stochastic gradient descent,'' \emph{The International Journal of Robotics
  Research}, vol.~35, no.~14, pp. 1717--1730, 2016.

\bibitem{ortiz_isdf_2022}
J.~Ortiz, A.~Clegg, J.~Dong, E.~Sucar, D.~Novotny, M.~Zollhoefer, and
  M.~Mukadam, ``isdf: Real-time neural signed distance fields for robot
  perception,'' \emph{arXiv preprint arXiv:2204.02296}, 2022.

\bibitem{lee_online_2019}
B.~Lee, C.~Zhang, Z.~Huang, and D.~D. Lee, ``Online continuous mapping using
  gaussian process implicit surfaces,'' in \emph{2019 International Conference
  on Robotics and Automation (ICRA)}.\hskip 1em plus 0.5em minus 0.4em\relax
  IEEE, 2019, pp. 6884--6890.

\bibitem{wu_faithful_2021}
L.~Wu, K.~M.~B. Lee, L.~Liu, and T.~Vidal-Calleja, ``Faithful euclidean
  distance field from log-gaussian process implicit surfaces,'' \emph{IEEE
  Robotics and Automation Letters}, vol.~6, no.~2, pp. 2461--2468, 2021.

\bibitem{liu_when_2020}
H.~Liu, Y.-S. Ong, X.~Shen, and J.~Cai, ``When gaussian process meets big data:
  A review of scalable gps,'' \emph{IEEE transactions on neural networks and
  learning systems}, vol.~31, no.~11, pp. 4405--4423, 2020.

\bibitem{kim_efficient_2022}
S.~Kim and J.~Kim, ``Efficient clustering for continuous occupancy mapping
  using a mixture of gaussian processes,'' \emph{Sensors}, vol.~22, no.~18, p.
  6832, 2022.

\bibitem{stork_ensemble_2020}
J.~A. Stork and T.~Stoyanov, ``Ensemble of sparse gaussian process experts for
  implicit surface mapping with streaming data,'' in \emph{2020 IEEE
  International Conference on Robotics and Automation (ICRA)}, 2020, pp.
  10\,758--10\,764.

\bibitem{tian_fast_2018}
Y.~Tian, K.~Wang, R.~Li, and L.~Zhao, ``A fast incremental map segmentation
  algorithm based on spectral clustering and quadtree,'' \emph{Advances in
  Mechanical Engineering}, vol.~10, no.~2, 2018.

\bibitem{bavle_s-graphs_2022}
H.~Bavle, J.~L. Sanchez-Lopez, M.~Shaheer, J.~Civera, and H.~Voos, ``S-graphs+:
  Real-time localization and mapping leveraging hierarchical representations,''
  \emph{arXiv preprint arXiv:2212.11770}, 2022.

\bibitem{ivan_online_2022}
J.-P.~A. Ivan, T.~Stoyanov, and J.~A. Stork, ``Online distance field priors for
  gaussian process implicit surfaces,'' \emph{IEEE Robotics and Automation
  Letters}, vol.~7, no.~4, pp. 8996--9003, 2022.

\bibitem{liu_adaptive_2022}
Y.~Liu, L.~Zhang, K.~Qian, L.~Sui, Y.~Lu, F.~Qian, T.~Yan, H.~Yu, and F.~Gao,
  ``An adaptive threshold line segment feature extraction algorithm for laser
  radar scanning environments,'' \emph{Electronics}, vol.~11, no.~11, p. 1759,
  2022.

\bibitem{gao_line_2018}
H.~Gao, X.~Zhang, Y.~Fang, and J.~Yuan, ``A line segment extraction algorithm
  using laser data based on seeded region growing,'' \emph{International
  Journal of Advanced Robotic Systems}, vol.~15, no.~1, 2018.

\bibitem{wen_cae-rlsm_2020}
J.~Wen, X.~Zhang, H.~Gao, J.~Yuan, and Y.~Fang, ``Cae-rlsm: Consistent and
  efficient redundant line segment merging for online feature map building,''
  \emph{IEEE Transactions on Instrumentation and Measurement}, vol.~69, no.~7,
  pp. 4222--4237, 2019.

\bibitem{von_luxburg_tutorial_2007}
U.~Von~Luxburg, ``A tutorial on spectral clustering,'' \emph{Statistics and
  computing}, vol.~17, pp. 395--416, 2007.

\bibitem{damle_simple_2019}
A.~Damle, V.~Minden, and L.~Ying, ``Simple, direct and efficient multi-way
  spectral clustering,'' \emph{Information and Inference: A Journal of the
  IMA}, vol.~8, no.~1, pp. 181--203, 2019.

\bibitem{Fiedler1973}
M.~Fiedler, ``Algebraic connectivity of graphs,'' \emph{Czechoslovak
  mathematical journal}, vol.~23, no.~2, pp. 298--305, 1973.

\bibitem{galy-fajou_adaptive_2021}
T.~Galy-Fajou and M.~Opper, ``Adaptive inducing points selection for gaussian
  processes,'' \emph{arXiv preprint arXiv:2107.10066}, 2021.

\end{thebibliography}

\end{document}